\documentclass[a4paper,conference]{IEEEtran}
\usepackage{amsmath,amssymb,amsfonts}
\DeclareMathOperator*{\argmax}{arg\,max}
\usepackage{algorithmic}
\usepackage{graphicx}
\usepackage{textcomp}
\usepackage{xcolor}
\usepackage{booktabs}
\usepackage{adjustbox}
\usepackage{gensymb}
\usepackage{subfig}

\newif\ifdebug
\debugtrue 

\usepackage{cite}

\ifCLASSINFOpdf
\else
\fi

\newcounter{MLCounter}

\begin{document}

\bstctlcite{IEEEexample:BSTcontrol}

\title{Respecting Domain Relations: \\ Hypothesis Invariance for Domain Generalization}


\author{\IEEEauthorblockN{Ziqi Wang}
\IEEEauthorblockA{
\textit{Delft University of Technology}\\
Delft, The Netherlands \\
z.wang-8@tudelft.nl}
\and
\IEEEauthorblockN{Marco Loog}
\IEEEauthorblockA{
\textit{Delft University of Technology}\\
Delft, The Netherlands \\
\textit{University of Copenhagen}\\
Copenhagen, Denmark}
\and
\IEEEauthorblockN{Jan van Gemert}
\IEEEauthorblockA{
\textit{Delft University of Technology}\\
Delft, The Netherlands}
}


%


\maketitle

\begin{abstract}
In domain generalization, multiple labeled non-independent and non-identically distributed source domains are available during training while neither the data nor the labels of target domains are. Currently, learning so-called \emph{domain invariant representations} (DIRs) is the prevalent approach to domain generalization. In this work, we define DIRs employed by existing works in probabilistic terms and show that by learning DIRs, overly strict requirements are imposed concerning the invariance. Particularly, DIRs aim to perfectly align representations of different domains, i.e. their input distributions. This is, however, not necessary for good generalization to a target domain and may even dispose of valuable classification information. We propose to learn so-called \emph{hypothesis invariant representations} (HIRs), which relax the invariance assumptions by merely aligning posteriors, instead of aligning representations. We report experimental results on public domain generalization datasets to show that learning HIRs is more effective than learning DIRs.  In fact, our approach can even compete with approaches using prior knowledge about domains.
\end{abstract}



%
\IEEEpeerreviewmaketitle

\section{Introduction}
A standard assumption for many machine learning algorithms is that training and test data are \emph{independent and identically distributed} (i.i.d.). In practice, however, training data may come from one source domain, while test data may come from another differently distributed target domain; violating the i.i.d. assumption~\cite{moreno2012unifying, quionero2009dataset, zadrozny2004learning, kouw2019review}. Domain generalization is the setting where we have access to labeled data from multiple source domains during training while neither the data nor the labels of the target domain are available \cite{ghifary2016scatter, li2017deeper, motiian2017unified, xu2014exploiting}.  The absence of target data makes it impossible to  infer distributional shifts between target and source domains, which significantly complicates the generalization to the target domain.



In some cases, strong prior knowledge about the relation between different domains that can be exploited in the building of a target classifier is available.  A well-known example is the artificially created rotated MNIST images \cite{ghifary2015domain}, where every domain---as the name eludes to---is a version of MNIST with all images rotated over a specific angle.  To make domain generalization broadly applicable, however, one cannot rely on such very specific prior information, which in most realistic cases, will simply not be available. Approaches that indeed do not rely on such prior knowledge often draw inspiration from the field of domain \emph{adaptation} \cite{daume2009frustratingly, kouw2019review} where learning domain invariant representations is prevalent. In domain adaptation (rather than domain generalization as considered in this work), the availability of input target data is assumed, which enables one to directly relate this input distribution to those of the various source domains.

Approaches that learn so-called domain invariant representations (DIRs) for domain adaptation inspire domain generalization. DIRs aim to remove those parts of the representation that are domain specific in an attempt to generalize better to the target domain \cite{ding2017deep, dinh2013fidos, ghifary2015domain,ghifary2016scatter,motiian2017unified}. In general supervised learning, to achieve invariance of representations, minimal information of the input data should be kept, which refers to the mutual information between the learned representations and the input \cite{achille2018emergence}. This, however, can compromise the sufficiency of representations for classification tasks if the learned invariance discards too much information. We transfer the definition for sufficiency of representations \cite{achille2018emergence} to the setting of domain generalization and define the invariance for domain generalization. We show that learning domain invariant representations is too strict for the invariance because DIRs force the source domains to be at the same location in the representation space, which we will show is unnecessary for generalizing to the target domain. So by learning DIRs, the relation between different domain distributions can be distorted. However, this relation between domains is useful in the way that it can be used to infer the target domain.

In this work, we propose to learn hypothesis invariant representations (HIRs) to relax the invariance requirement of representation learning Where DIRs align data sample representations between domains, our proposed HIRs merely align classifier predictions between domains. If there is any useful relation between domains, like the rotation in rotated MNIST, learning hypothesis invariant representations differs from DIRs in that HIRs can preserve the relative location of domains in the representation space without having prior domain knowledge.  As it turns out, HIRs can even compete with approaches using prior knowledge. We demonstrate that hypothesis invariant representations can solve prior shifts as well.  This setting is usually not considered by domain invariant representations. Moreover, when the network is trained on augmented data \cite{hendrycks2018benchmarking}, we show that learning hypothesis invariant representations can improve the performance over the baseline where various types of corruptions are aggregated and trained only with a classification loss.

All in all, this work makes the following contributions: 
\begin{itemize}
    \item We introduce the notions of sufficiency and invariance of representations to domain generalization.
    
    \item We present a probabilistic formulation to categorize and evaluate DIRs in terms of sufficiency and invariance.
    
    \item We introduce hypothesis invariant representations (HIRs). HIRs relax the invariance demands in DIRs by allowing to preserve useful relations in representation space while aligning predictions instead of representations.
    
    \item We  compare HIRs to DIRs and against using prior knowledge about the domain distributions.

\end{itemize}

\section{Related work}\label{section: related work}

A similar setting to domain generalization is domain adaptation \cite{daume2009frustratingly, kouw2019review}, where both the data and label of the source domain are available while only the data from the target domain is accessible during training. When multiple labeled source domains and unlabeled target domains are available, it is referred to as multi-source domain adaptation~\cite{daume2009frustratingly,dredze2008online,hoffman2012discovering,mansour2009domain,zhang2015multi}. For both domain adaptation and domain generalization, the challenge is the distribution shift between source domains and target domains. Three types of distribution shifts are usually involved: covariate shift \cite{shimodaira2000improving}, concept drift  \cite{widmer1996learning}, and prior shift \cite{webb2005application}. We refer the reader to the study in \cite{moreno2012unifying} for a comprehensive treatment of these three shifts. In this work, both covariate shift and prior shift are considered. 

Some domain generalization approaches can exploit strong prior knowledge of domains to infer the target domain distribution. LG \cite{shankar2018generalizing} assumes a continuous representation space for domains with a specific order and applies perturbations on training domains to generalize to the unseen domain. DIVA \cite{ilse2019diva} designs a generative model to decompose the representations of domain, class and variations. Due to the generative function of the network, it is possible to simulate unseen domains, especially if the domain is generated in an order. The disadvantage of these approaches is that the prior knowledge of domains is not always available. HIR learning does not require prior knowledge of domains.




When prior knowledge is not available, several approaches draw inspiration from  domain adaptation to learn domain invariant representations for domain generalization. Guided by theoretical proof \cite{ben2007analysis,ben2010theory,ben2010impossibility}, learning domain invariant representations for domain adaptation by aligning distributions of source domains and target domains is prevalent \cite{ganin2016domain, fernando2013unsupervised, bhushan2018deepjdot,courty2017joint}. KL divergence \cite{yu2013kl} and JS divergence \cite{endres2003new} are used to measure the divergence between distributions. In this work, we formulate domain invariant representation in a probabilistic setting and extend the discussion of invariance and sufficiency \cite{achille2018emergence} into domain generalization. We show that there is a  trade-off between sufficiency and invariance learning, which aims to keep minimal information of domains.


Learning a domain invariant representation (DIR) for domain generalization can be achieved with kernel machines, e.g., DICA \cite{muandet2013domain}, MDA \cite{hu2019domain}, SCA \cite{ghifary2016scatter} and deep learning \cite{ding2017deep, ghifary2015domain, li2018domain, motiian2017unified}. DA \cite{ganin2016domain} uses a gradient reverse layer for the domain classifier prevent the network from distinguishing different domains. HEX \cite{wang2019learning} proposes to separate the domain specific representations of different domains from the domain invariant representations of all possible domains. D-MTAE \cite{ghifary2015domain} designs a multi-task denoising autoencoder to reconstruct each domain with the goal to learn DIRs that are robust to noise. CIDDG \cite{li2018deep} designs a new architecture to solve the problem of prior shift in domain generalization. MMD-AAE \cite{li2018domain} applies \emph{Maximum Mean Discrepancy (MMD)} \cite{borgwardt2006integrating} loss in the latent space of an encoder-decoder network to align the representations of different domains. CCSA \cite{motiian2017unified} aligns the representations of different domains by minimizing the Euclidean distance in the feature space. DBADG \cite{xu2014exploiting} introduces a low-rank constraint to the weights of the network to not learn domain specific information. Different from learning DIRs, our hypothesis invariant representation learning does not strictly align domains in the latent space but rather keeps the relative positions between domains. 


Meta-learning for domain generalization introduces a way to train a more robust model. MLDG \cite{li2018learning} introduces meta-learning into the training procedure by treating part of the training data as from the target domain to learn a model that can generalize well to the unseen domain. Epi-FCR \cite{li2019episodic} proposes to learn a domain agnostic model by shuffling the feature extractors of different domains. We also compare our approach to meta-learning.


\section{Method}\label{section: method}
We first give a probabilistic analysis of domain generalization. The analysis details sufficiency and invariance to show the trade-off between DIR learning and sufficiency of representations. Existing DIR approaches can be categorized in two conditions: class-agnostic and class-conditional. We formulate probabilistic definitions for DIRs under these two conditions and define hypothesis invariant representations to compare DIRs and HIRs. We show that (1) learning HIRs is less strict than learning DIRs; (2)  HIRs can tackle prior shift and (3) a loss to learn HIRs.

\subsection{Preliminaries}\label{section: preliminaries}
Inspired by \cite{xie2017controllable} and  \cite{akuzawa2019adversarial}, we examine the relationship between distributions of domains and classes. We formalize domain generalization in the setting of classification. The label space and the domain space are denoted as $\mathcal{Y}$ and $\mathcal{D}$ respectively where class label $Y$ and domain $D$ are sampled from distributions $P_Y$ and $P_D$. Seen source domains $D_{s} \in \mathcal{D}$ and unseen target domains $D_{t} \in \mathcal{D}$ are all in the space $\mathcal{D}$. The distribution of $X$ is in input space $\mathcal{X} = \mathbb{R}^d$ conditioned on $\mathcal{Y} \times \mathcal{D}$ and is denoted as $P_{X|Y,D}$. For simplicity, we consider discrete domains and labels. 

We introduce a latent space $\mathcal{Z}$ for representations to facilitate the discussion about DIRs and HIRs later. We denote the mapping function from the input space to the latent space as a \emph{representation function} $r: \mathcal{X} \xrightarrow{} \mathcal{Z}$ and a \emph{hypothesis function} from the latent space to the predicted labels $h: \mathcal{Z} \xrightarrow{} \mathcal{\hat{Y}}$. 

\begin{figure}
    \centering
    \includegraphics[width =0.35\textwidth]{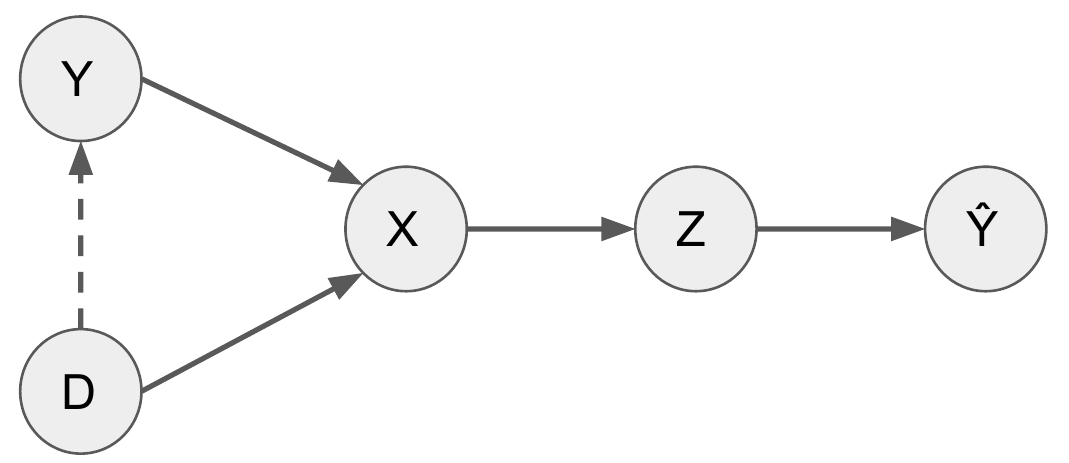}
    \caption{The graphical model shows the relationship from domain $D$, class $Y$ to the input data $X$. A latent representation $Z$ is learned to predict $\hat{Y}$. If $Y$ is independent of $D$, $P_{Y|D} = P_Y$.  Else if $Y$ is dependent on $D$, $P_{Y|D} \neq P_Y$, which is denoted as an arrow pointing from $D$ to $Y$. This dependency causes prior shift.}
    \label{fig:independent}
\end{figure}

Prior shift is caused by the unbalance of classes across domains, $P_{Y|D} \neq P_Y$. If the prior of the target domain $P_{Y|D_t}$ varies significantly from the prior of the source domain $P_{Y|D_s}$, this prior shift will lead to the failure of an approach which does not consider this type of distribution shift. To cover the possible prior shift in domain generalization, we set our graphical model in two different conditions: $Y$ is independent of $D$ or not. If the distribution of $Y$ is independent of domains, $P_{Y|D} = P_Y$, it means there is no prior shift as shown in the graphical model in Fig.~\ref{fig:independent}. Else if $Y$ is dependent on $D$, $P_{Y|D} \neq P_Y$, prior shift exists.

\subsection{Sufficiency of representations}\label{section: Sufficiency}
According to the work of Achille and Soatto \cite{achille2018emergence}, a representation $Z$ is considered as sufficient for the classification task if $P_{Y|X,Z} = P_{Y|Z}$ in a Markov chain $X \xrightarrow{} Z \xrightarrow{} Y$ for general supervised machine learning setting. That implies the mutual information between the input $X$ and the label $Y$ equals the mutual information between the latent representation $Z$ and the label $Y$, $I(X;Y) = I(Z;Y)$ because the representation $Z$ distills all the useful information for the classification task from the input $X$. Here we extend the definition of sufficiency into domain generalization. If we assume that the representation of each domain is sufficient, the posterior can be defined to satisfy: 
\begin{equation}
\label{eq: sufficiency}
\begin{aligned}
    & P_{Y|X,Z,D} = P_{Y|Z,D},\\
    & \forall D \in \mathcal{D}.
\end{aligned}
\end{equation}

\subsection{Invariance of representations}\label{section: Invariance}
A concept that recurs in many domain generalization works \cite{akuzawa2019adversarial, li2018domain, muandet2013domain}, is the so-called domain invariant representations which are supposed to be the essential representations of all domains and invariant to different domains, not only source domains but also the target domain. However, many approaches aim to learn domain invariant representations without clear definitions. Therefore we formulate two mostly adopted conditions of DIRs as below. 

\textbf{Class-agnostic DIRs} 
refer to representations $Z$ that are independent of the domain $D$, irrespective of labels $Y$. We formulate it as:
\begin{equation}
\label{eq: MMD}
\begin{aligned}
    & P_{Z|D} = P_Z,\\
    & \forall D \in \mathcal{D}.\\
\end{aligned}
\end{equation}

Approaches that fall in this category align representations from multiple source domains without considering the labels.

\textbf{Class-conditional DIRs} 
are conditioned on both the domains $D$ and the labels $Y$, so it is referred to as class-conditional DIR:
\begin{equation}
\label{eq: cc dir}
\begin{aligned}
    & P_{Z|Y,D} = P_{Z|Y},\\
    & \forall D \in \mathcal{D}.
\end{aligned}
\end{equation}
Different from class-agnostic DIRs, approaches that aim to learn class-conditional DIRs align representations of the same class but different domains. Some works adopt both class-agnostic and class-conditional domain invariant representation learning. We sort existing approaches in Table~\ref{tab: categorization}.

\begin{table}[bt]
\centering
\caption{Categorization of approaches. Existing approaches that learn domain invariant representations for domain generalization are sorted according to the definitions \eqref{eq: MMD}, \eqref{eq: cc dir}.}
\label{tab: categorization}
\begin{adjustbox}{width=0.9\columnwidth,center}
\begin{tabular}{@{}cccc@{}}
\toprule
&Class-agnostic DIRs &Class-conditional DIRs &both \\\cmidrule(r){2-4}
 &DA \cite{ganin2016domain} &CCSA \cite{motiian2017unified} &SCA \cite{ghifary2016scatter}\\
 
&MMD-AAE \cite{li2018domain} & CIDDG \cite{li2018deep} & MDA \cite{hu2019domain}\\
 
&D-MTAE \cite{ghifary2015domain} && DICA \cite{muandet2013domain}\\

&HEX \cite{wang2019learning}\\
\bottomrule
\end{tabular}
\end{adjustbox}
\end{table}


Learning domain invariant representations adds constraints on the representation invariance. However, if the algorithm focuses on learning DIRs, the useful domain specific information in the input $X$ may be discarded so the sufficiency is compromised. If a method is forced to learn DIRs, a degenerate example of a trivial representation can be a vector full of zeros, which is invariant to all domains or hypotheses, but, such a vector is not sufficient for the classification task. Furthermore, by the definitions of DIRs, all domains are aligned with each other, and the relation between domains is no longer available. Thus, if the domains are sampled according to a specific order, this order information is subsequently removed by learning DIRs. Instead, we propose to learn HIRs to relax the constraints on the invariance so the relation between domains is retained.

\textbf{Hypothesis invariant representations} make domains invariant to the \emph{prediction hypotheses} instead of DIRs that make the \emph{feature representations} invariant to domains. Thus, the aim of HIRs is to align the predictions for representations from different domains, where a prediction label for two domains $a,b$ is aligned with the hypothesis function: $\argmax_Y P_{Y|Z,D=a} = \argmax_Y P_{Y|Z,D=b}$. The requirement of DIRs for aligning feature representations from different domains is too strict, as only the prediction hypothesis needs to be domain invariant.  HIRs should satisfy:
\begin{equation}
\label{eq: domain_invariant}
\begin{aligned}
   &\argmax_Y P_{Y|Z,D} = \argmax_Y P_{Y|Z},\\
   & \forall D \in \mathcal{D}.
\end{aligned}
\end{equation}

\begin{figure}
    \centering
    \includegraphics[width = 0.35\textwidth]{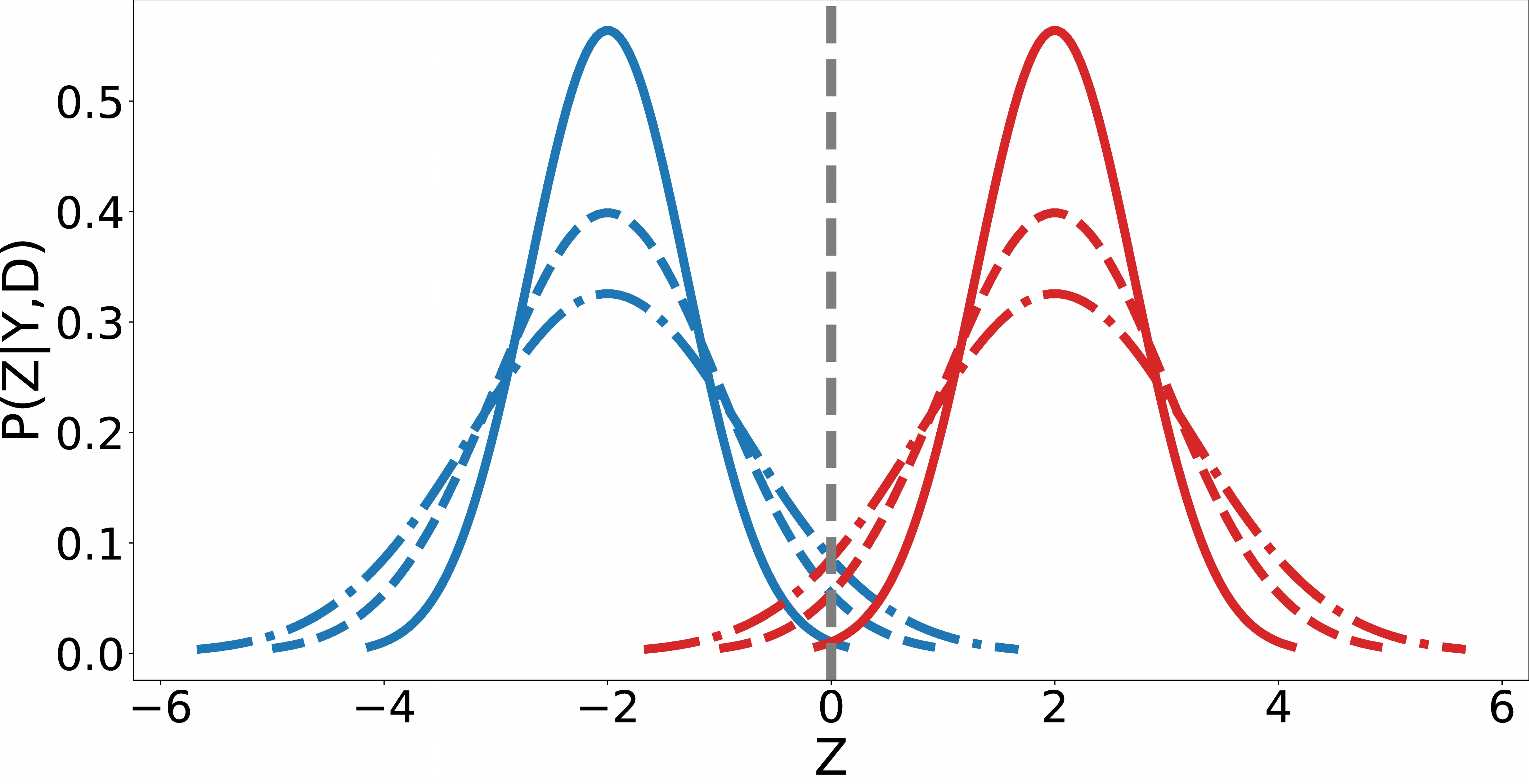}
    \caption{Representations of three domains. Colors represent the two classes of the three domains. This shows that learning DIRs is too strict compared to learning HIRs because representations $Z$ of the three domains are hypothesis invariant as in \eqref{eq: domain_invariant} but not domain invariant as in \eqref{eq: MMD} and \eqref{eq: cc dir}. A threshold function on $Z$ can serve as a hypothesis function $h$ with low error for the binary classification task on all the three domains.}
    \label{fig:strict}
\end{figure}

What matters for the final classification result is not DIRs, as in \eqref{eq: MMD} or \eqref{eq: cc dir}, but $\argmax_Y P_{Y|Z,D}$, which is more relaxed because it can still be satisfied even if the representations differ across domains. See for example Fig.~\ref{fig:strict} where neither the distributions of representations $P_{Z|Y,D}$ nor the posterior $P_{Y|Z,D}$ is domain invariant for all the domains presented, but the hypothesis function $h$ can give labels that satisfy \eqref{eq: domain_invariant}. So for HIR, $P_{Z|Y,D=a} \neq P_{Z|Y,D=b}$ can hold.

Note that \eqref{eq: domain_invariant} in itself does not imply correct classification performance but only guarantees that the samples of all domains share the same prediction hypothesis function $h: \mathcal{Z} \xrightarrow{} \mathcal{\hat{Y}}$, which can be completely different from the ground truth hypothesis function. If the hypothesis function $h$ is inappropriate, then the samples could all be wrongly classified. The classification result is related to the sufficiency of the representation. So in practice, both the HIR learning and the classification learning are required.

\subsection{HIRs and DIRs comparison}
To further examine the relationship between HIRs, class-agnostic DIRs and class-conditional DIRs, we expand the posterior $P_{Y|Z,D}$ as: 
\begin{equation}
\label{eq: posterior invariant_dependent}
    P_{Y|Z,D} = \frac{P_{Z|Y,D}P_{Y|D}}{P_{Z|D}}.
\end{equation}
If each of the three items on the right-hand side of \eqref{eq: posterior invariant_dependent} is independent of $D$, we will get class-agnostic DIRs as in \eqref{eq: MMD}, class-conditional DIRs as in \eqref{eq: cc dir}, and domain invariant priors separately. If all the three items are independent of domain, then the posterior $P_{Y|Z,D}$ is domain invariant by construction, that is, $P_{Y|Z,D} = P_{Y|Z}$. So if there is no prior shift, $P_{Y|D} = P_{Y}$, the DIR is sufficient for the HIR. To the contrary, if $P_{Y|Z,D} = P_{Y|Z}$ holds, DIRs cannot be guaranteed. This expansion shows first, learning DIRs cannot tackle the prior shift when $Y$ is dependent on $D$, and second, DIRs are sufficient but not necessary for HIRs. Therefore, learning HIRs is a more relaxed regularization on invariance and can align the priors of different domains.



\begin{figure*}
\centering
\includegraphics[width=0.75\textwidth]{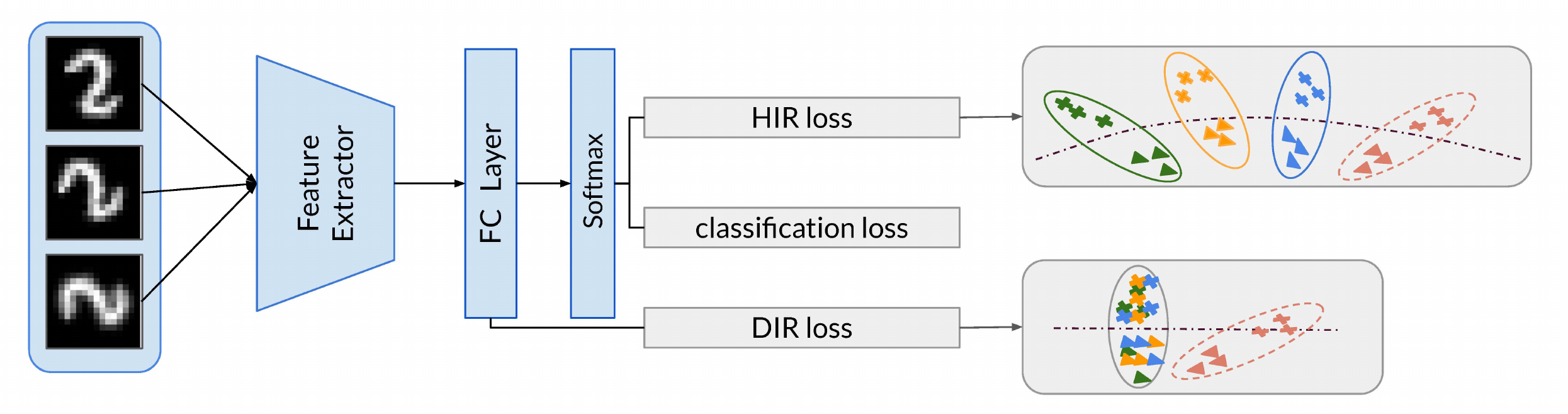}
\caption{Comparison between HIR and DIR losses. The inputs are three domains of digits with different rotated angles. In the latent space, different colors represent domains and classes are marked by different shapes. The dash-line circles the unseen target domain. For learning DIRs, the loss is applied after the feature extractor which aims to align the representations of samples from different domains. HIR loss is applied after the Softmax layer to align the distributions of posteriors from different domains. With HIR loss, representations may keep the global structure without being strictly aligned.}
\label{fig: network}
\end{figure*}

\begin{figure}[bt]
    \centering
  \subfloat[KL asymmetric\label{subfig: kl_half}]{%
       \includegraphics[width=0.35\linewidth]{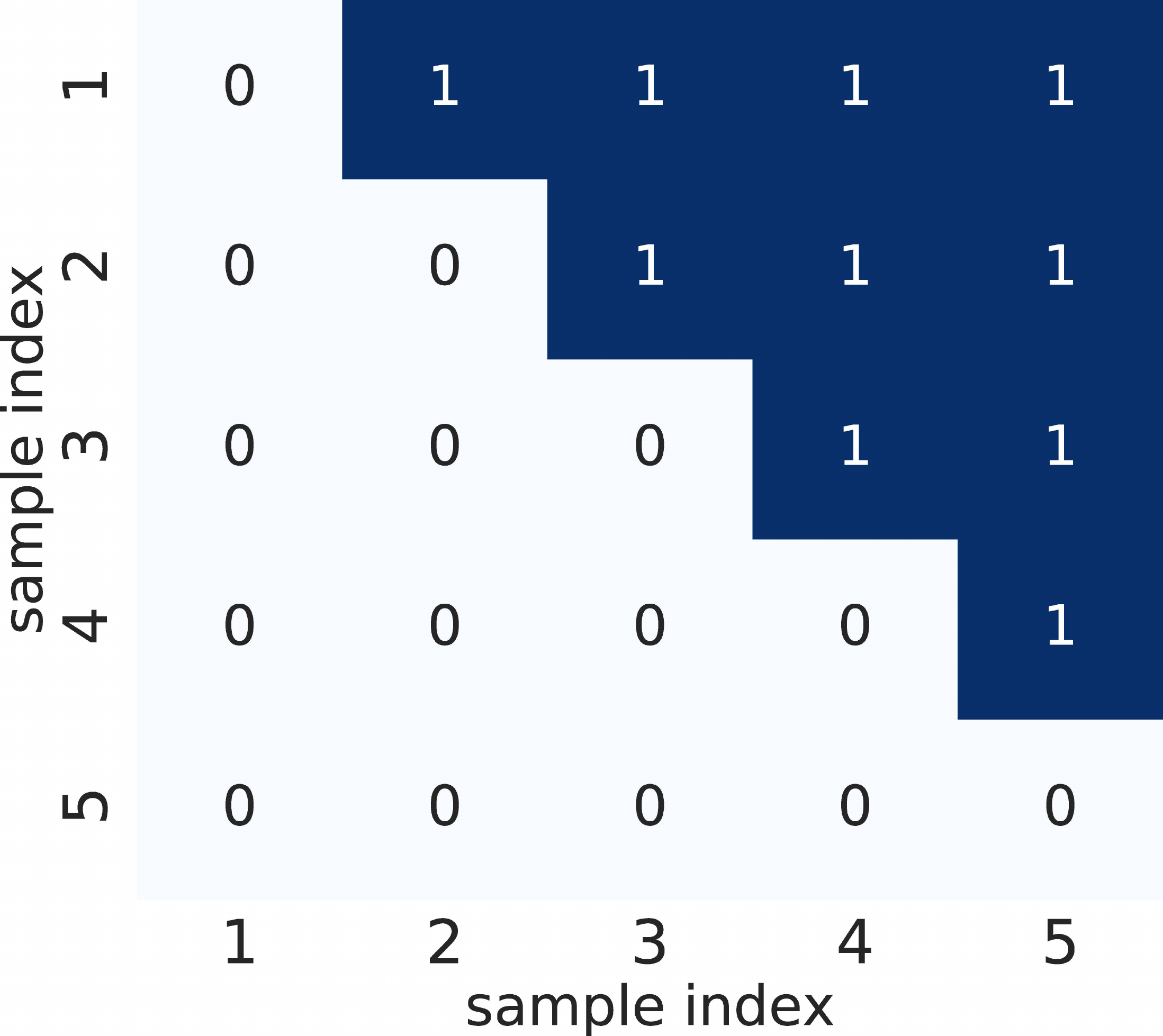}}
       \vspace{0.1cm}
  \subfloat[KL symmetric\label{subfig: kl_both}]{%
        \includegraphics[width=0.35\linewidth]{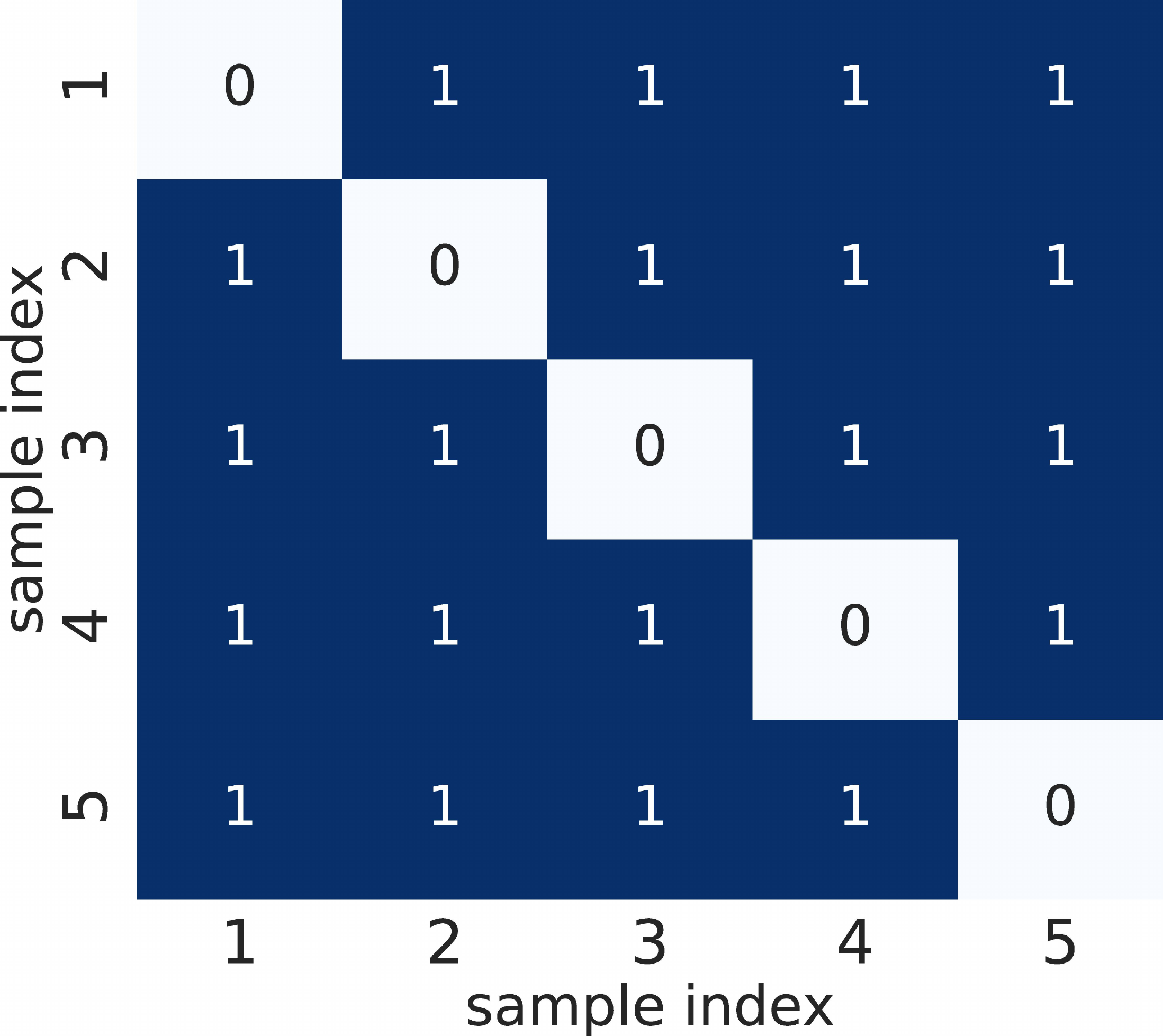}}
    \caption{Asymmetric and symmetric KL divergence between 5 samples. The two matrices show all the possible matches among 5 samples. If one match is denoted as 1, then the KL divergence is computed, otherwise not. We use the asymmetric KL divergence in the HIR loss.}
  \label{fig: kl divergence} 
\end{figure}

\subsection{Aligning hypotheses: The HIR loss}
Learning domain invariant posteriors is an approximation to learn HIRs, as in \eqref{eq: domain_invariant}, because it regularizes the invariance aspect of posteriors to generalize to the unseen target domain. Based on the analyses above on \eqref{eq: posterior invariant_dependent}, learning HIRs by aligning the posteriors of domains $P_{Y|Z,D}$ can tackle the prior shift in practice and it is a more relaxed invariance regularization for representation learning. However, the distribution $P_{Y|Z,D}$ is usually not available. To avoid arbitrary density estimation of $P_{Y|Z,D}$ and guide the network to learn HIRs, we propose to align the domain-agnostic class-conditional posteriors of training data by minimizing the KL divergence as below:
\begin{equation}\label{eq: kl}
L_h = \sum_{i=1}^n \sum_{j=i+1}^n \sum_{c=1}^m P(\hat{Y}_i|Z_i, Y_i=c) \cdot \log(\frac{P(\hat{Y}_i|Z_i, Y_i=c)}{P(\hat{Y}_j|Z_j, Y_j=c)}),
\end{equation}{}
where $n$ is the number of samples for class $c$, $i$ and $j$ are indices of samples. The KL divergence is computed for all $m$ classes separately and summed up. The comparison between HIR loss and general DIR loss is presented in Fig.~\ref{fig: network}.

We choose KL divergence to align the posteriors because it is a measurement for the difference between two distributions, but not only a distance between distributions. KL divergence is asymmetric and in this work we only calculate one direction between two samples. That is because there is no target distribution in our case. Instead, posteriors of both the two samples can change during training. So calculating the symmetric version of KL divergence or JS divergence is not very different from calculating the asymmetric KL divergence in our setting. The difference between the two approaches is presented in Fig.~\ref{fig: kl divergence}.

The sufficiency of representations is guaranteed by the cross-entropy loss which is computed as: 
\begin{equation}\label{eq: crossentropy}
    L_c = -\frac{1}{n}\sum_i^n\sum_c^m Y_i \cdot \log(P(\hat{Y_i}=c|Z_i)).
\end{equation}{}
The trade-off between the sufficiency and the invariance learning is regulated by a coefficient $\alpha$:
\begin{equation}\label{eq: cross+kl}
    L = L_c + \alpha \cdot L_h,
\end{equation}{}
where $\alpha$ is a tunable parameter during training. Different $\alpha$ values should be selected according to the scale of the HIR loss $L_h$ to match the scale of $L_c$.

~\\

\section{Experiments}\label{section: experiments}
We show empirically that our approach respects the relations between domains without using prior knowledge of domains because it does not align the representations as strict as learning DIRs. We also show that our approach can do well on datasets which lack an obvious relation or global structure among domains. We compare our results with other existing approaches, especially the approaches that focus on learning DIRs and the approaches that can exploit prior knowledge of domain distributions. In addition, we also demonstrate the effectiveness of HIR learning on the data augmentation task. The hold-one-domain-out domain generalization setting is adopted for all experiments, that is, neither the label nor the data of the test domain is available during training. The trained network is applied on the unseen domain for evaluation without any adaptation or fine tuning. 
\subsection{Datasets}
We examine the results of learning HIRs on three datasets, (1) rotated MNIST dataset with clear prior knowledge about the global structure of domains, (2) VLCS where there is no order for domains and (3) tiny ImageNet-C which consists of 7 types of corruptions, where each corruption is a domain.

\subsubsection{\textbf{Rotated MNIST}}
Rotated MNIST dataset consists of 6 domains with the original domain $\mathcal{M}_{0\degree}$ and it rotated by ${15\degree}$, ${30\degree}$, ${45\degree}$, ${60\degree}$ and ${75\degree}$. Each domain has 10 classes of hand written digits from 0 to 9, and 100 images for each class. This dataset has a specific rotation order for domains so it is usually used to test approaches where prior domain information is used.
\subsubsection{\textbf{VLCS}}
VLCS dataset \cite{fang2013unbiased} has four domains, each domain is a different dataset collected under different backgrounds, namely PASCAL VOC2007 (V) \cite{everingham2010pascal}, LabelMe (L) \cite{russell2008labelme}, Caltech-101 (C) \cite{fei2004learning} and SUN09 (S) \cite{choi2010exploiting} with 5 common classes, bird, car, chair, dog and person. To be consistent with other approaches \cite{ghifary2015domain, li2019episodic, motiian2017unified}, we also use DeCAF features in the experiments. The domains in VLCS do not follow an obvious order, so the approaches using prior knowledge cannot be applied on this dataset.

\begin{table*}[t!]
\centering
\caption{Results on rotated MNIST dataset. Rotated MNIST is a dataset with a global structure for the domains, where the domains are rotated by a fixed angle. AGG is the baseline setting with only a classification loss without using HIR loss.  We show that on the ordered dataset, approaches using prior knowledge perform the best. Moreover, our HIR learning can compete with these approaches without using the prior knowledge.}
\label{tab: rmnist}
\begin{tabular}{@{}cccccccccc@{}}
\toprule
&&\textbf{Methods} &$\mathcal{M}_{0\degree}$  &$\mathcal{M}_{15\degree}$ &$\mathcal{M}_{30\degree}$ &$\mathcal{M}_{45\degree}$ &$\mathcal{M}_{60\degree}$ &$\mathcal{M}_{75\degree}$ &Avg.\\ \cmidrule(r){3-10}
&prior &LG &89.7 &97.8 &98.0 &97.1 &96.6 &92.1 &95.3\\
&knowledge&DIVA&93.5 &99.3 &99.1 &99.2 &99.3 &93.0 &\textbf{97.2}\\  \cmidrule(r){2-10}
& no &D-MTAE &82.5 &96.3 &93.4 &78.6 &94.2 &80.5 &87.5\\
&prior&CCSA &84.6 &95.6 &94.6 &82.9 &94.8 &82.1 &89.1\\
&knowledge &MMD-AAE &83.7 &96.9 &95.7 &85.2 &95.9 &81.2 &89.8\\
&&DA &86.7 &98.0 &97.8 &97.4 &96.9 &89.1 &94.3\\
&&HEX&90.1 &98.9 &98.9 &98.8 &98.3 &90.0 &95.8\\ \cmidrule(r){3-10}
&&AGG &89.87 &99.41 &98.98 &95.14 &98.63 &91.13 &95.53\\
&Ours &HIR &90.34 $\pm$0.88 &99.75 $\pm$0.18 &99.40 $\pm$0.21 &96.17 $\pm$0.71 &99.25 $\pm$0.26 &91.26 $\pm$0.66 &\textbf{96.03}\\  
\bottomrule
\end{tabular}
\end{table*}


\subsubsection{\textbf{Tiny ImageNet-C}}
The Tiny ImageNet dataset is a subset of ImageNet with 200 selected classes and 500 images per class. Tiny ImageNet-C has 7 domains, where each domain is one type of corruption of the original Tiny ImageNet dataset. The 7 types of corruptions, Gaussian noise, Impulse noise, JPEG compression, Defocus blur, Motion blur, Zoom blur and Glass blur are selected by us. We deployed the corruption methods from \cite{hendrycks2018benchmarking} and used the severest level 5 corruption. We designed this dataset to evaluate the effectiveness of HIR learning on data augmentation task.

\subsection{Results}
\subsubsection{\textbf{Rotated MNIST}}
As we expect that there is a global structure for the order of rotated angles, as shown in Fig.~\ref{fig: network}, the decision boundaries of domains in the middle of this global structure can be interpolated by the domains from both sides, e.g., $\mathcal{M}_{15\degree}$ can be inferred from $\mathcal{M}_{0\degree}$ and $\mathcal{M}_{30\degree}$. For the same reason, domain $\mathcal{M}_{0\degree}$ and $\mathcal{M}_{75\degree}$ are significantly more difficult to be generalized to, compared to the other domains. These two domains are located at the two ends of this global structure of all the domains, so the decision boundary can only be inferred from all the domains at only one side of the global structure. 

We adopt the same network architecture of CCSA \cite{motiian2017unified}, which has two convolutional layers with 32 kernels each and three fully connected layers. We report the average results of 10 repetitions for both the aggregation training setting (AGG) and the HIR setting in Table~\ref{tab: rmnist}. For AGG, the network is trained with only \eqref{eq: crossentropy} and no domain information is used as a baseline.  For HIR the  KL divergence is regularized as in \eqref{eq: cross+kl}. For HIR, a batch size of 250 is used, with 5 samples from each class and each domain. We use Adam for optimization with a learning rate of 1e-3. The coefficient $\alpha$ is set to be 1e-3. We can see that AGG with only classification loss can already give much better results compared to CCSA with the same architecture. After imposing the HIR loss, the performance can be further improved. We compare our HIR learning with approaches using prior knowledge and DIR learning without using prior knowledge. The results show that HIR learning can achieve better results than DIR learning and can compete with methods using prior knowledge, despite the fact that prior knowledge is unavailable in HIR learning.

\begin{table}[bt]
    \centering
    \caption{Pair/unpair experiment results on rotated MNIST. The AGG setting uses only the  classification loss and gives similar performance for both the paired and unpaired inputs. Our HIR learning is more effective on unpaired inputs for this dataset.}
    \label{tab:pair_rmnist}
    \begin{adjustbox}{width=\columnwidth,center}
    \begin{tabular}{@{}cccccc@{}}
    \toprule
         &\textbf{Domains} &unpaired AGG &paired AGG &unpaired HIR &paired HIR\\\cmidrule(r){3-6}
         &$\mathcal{M}_{0\degree}$ &45.83$\pm$2.67 &44.41$\pm$2.86 &79.99$\pm$3.78 &57.30$\pm$4.05\\
         &$\mathcal{M}_{15\degree}$ &65.83 $\pm$ 3.08 &66.33 $\pm$ 3.47&94.83 $\pm$ 4.14 &70.44 $\pm$ 4.55\\
         &$\mathcal{M}_{30\degree}$ &71.30 $\pm$ 4.86 &70.50 $\pm$ 4.56 &94.32 $\pm$ 4.07 &72.74 $\pm$ 3.65\\
         &$\mathcal{M}_{45\degree}$ &63.76 $\pm$ 3.94 &64.02 $\pm$ 3.51 &85.54 $\pm$ 5.34&63.63 $\pm$ 1.96 \\
         &$\mathcal{M}_{60\degree}$ &60.37 $\pm$ 3.05 &62.46 $\pm$ 4.94 &89.62 $\pm$ 4.57 &67.63 $\pm$ 4.91\\
         &$\mathcal{M}_{75\degree}$  &44.91 $\pm$ 2.65 &44.46 $\pm$ 3.30 &76.37 $\pm$ 5.64  &53.51 $\pm$ 2.49\\
         &Avg. &58.67 &58.70 &86.78 &64.21\\
    \bottomrule
    \end{tabular}
    \end{adjustbox}
\end{table}

\begin{figure}[tb]
    \centering
    \includegraphics[width=0.35\textwidth]{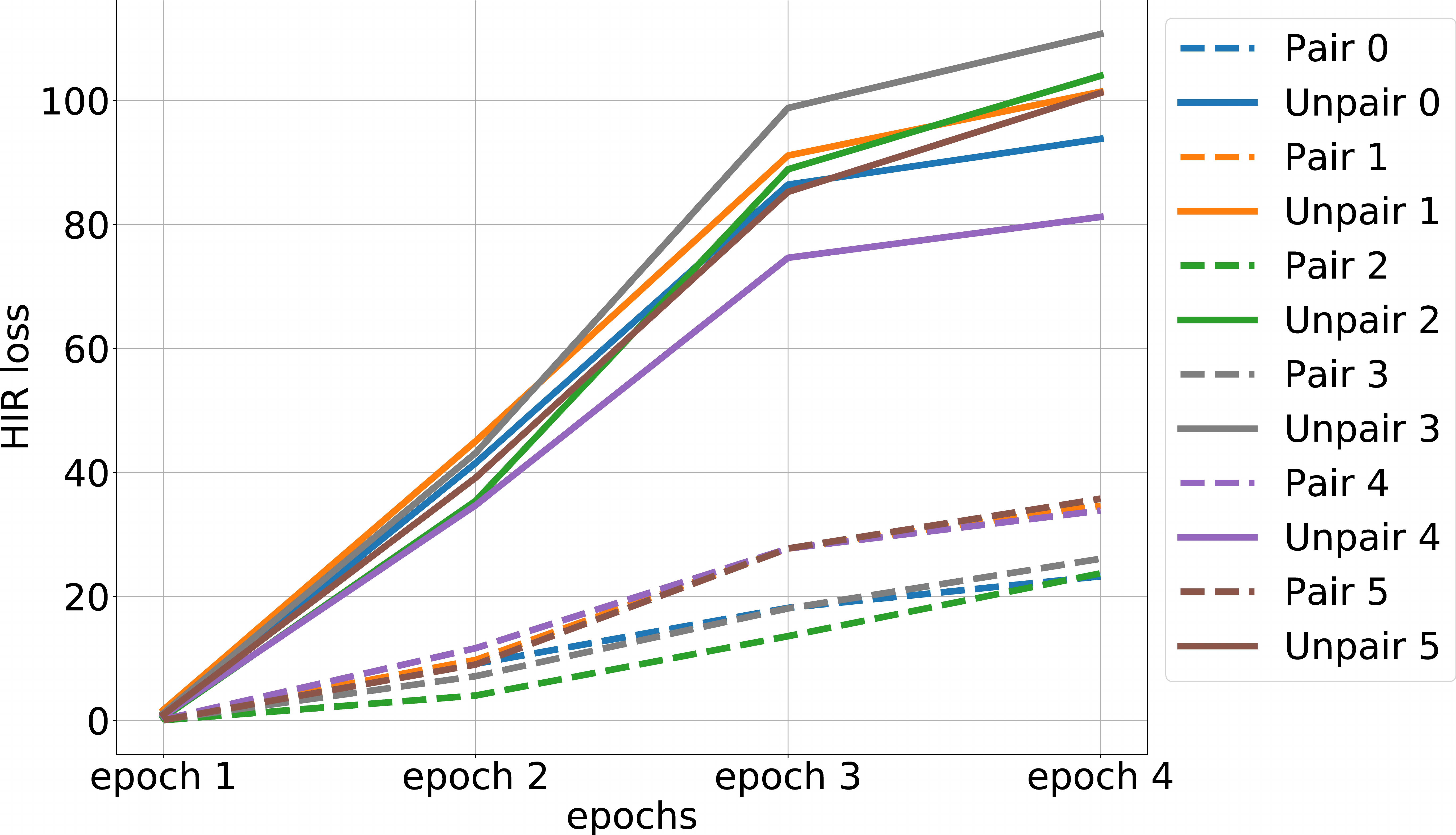}
    \caption{HIR losses of all 6 domains for the pair/unpair experiment on rotated MNIST dataset. The scale of the unpaired HIR loss is much larger than that of the paired inputs. So there is more room for HIR loss to contribute to the posterior alignment. For the paired inputs, the posteriors are similar within the pair so the HIR loss is low. Further regularizing HIR loss will lead the network to overfit to each pair of images.}
    \label{fig:kl_losses}
\end{figure}

\begin{table*}[ht]
\centering
\caption{Results on VLCS dataset. The domains in VLCS dataset do not follow a specific order or distribution, so prior knowledge cannot be used on this dataset. AGG is the baseline setting with only a classification loss without HIR loss. Our HIR learning performs better than the approaches that learn domain invariant representations.}
\label{tab: vlcs}
\begin{tabular}{@{}lcccccccccc@{}}
\toprule
&\textbf{Domains} &D-MTAE &CIDDG &DBADG &MMD-AAE &MLDG &Epi-FCR  &CCSA &\textbf{AGG} &\textbf{HIR}\\ \cmidrule(r){3-11}
&V    &63.90 &64.38 &69.99 &67.70 &67.7 &67.1 &67.10 &65.4 &69.10 $\pm$ 1.8\\
&L    &60.13 &63.06 &63.49 &62.60 &61.3 &64.3 &62.10 &60.6 &62.22 $\pm$ 1.7\\
&C    &89.05 &88.83 &93.63 &94.40 &94.4 &94.1 &92.30 &93.1 &95.39 $\pm$ 0.9\\
&S    &61.33 &62.10 &61.32 &64.40 &65.9 &65.9 &59.10 &65.8 &65.71 $\pm$ 1.6\\
&Avg  &68.60 &69.59 &72.11 &72.28 &72.3 &72.9 &70.15 &71.2 &\textbf{73.10}\\
\bottomrule
\end{tabular}
\end{table*}

\begin{figure*}[ht] 
    \centering
  \subfloat[uncorrupted\label{subfig: original}]{%
       \includegraphics[width=0.11\linewidth]{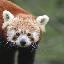}}
  \subfloat[Gauss.~noise\label{subfig: gaussina_noise}]{%
        \includegraphics[width=0.11\linewidth]{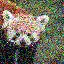}}
  \subfloat[Impulse noise\label{subfig: impulse_noise}]{%
        \includegraphics[width=0.11\linewidth]{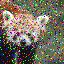}}
    \subfloat[JPEG\label{subfig: jpeg_compression}]{%
        \includegraphics[width=0.11\linewidth]{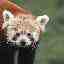}}
  \subfloat[Defocus blur\label{subfig: defocus_blur}]{%
       \includegraphics[width=0.11\linewidth]{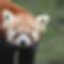}}
  \subfloat[Motion blur\label{subfig: motion_blur}]{%
        \includegraphics[width=0.11\linewidth]{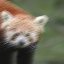}}
  \subfloat[Zoom blur\label{subfig: zoom_blur}]{%
        \includegraphics[width=0.11\linewidth]{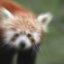}}
  \subfloat[Glass blur\label{subfig: glass_blur}]{%
        \includegraphics[width=0.11\linewidth]{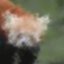}}
  \caption{Corruptions in Tiny ImageNet-C dataset. All images are corrupted at the highest severity \cite{hendrycks2018benchmarking}.}
  \label{fig: corruptions} 
\end{figure*}


\begin{figure*}[ht]
    \centering
    \includegraphics[width=.8\textwidth]{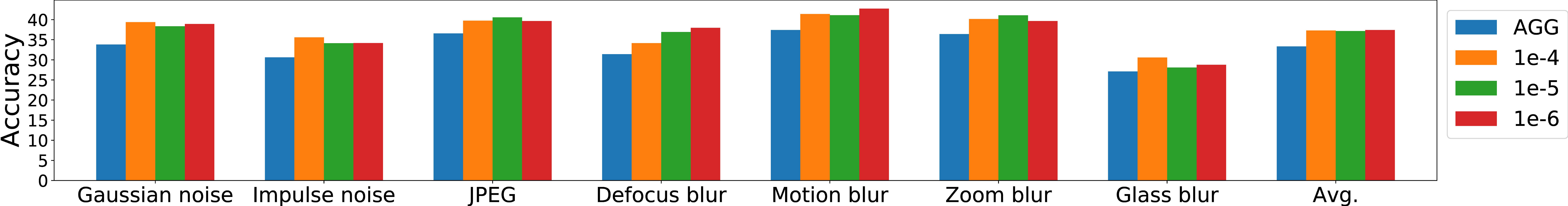}
    \caption{Results on Tiny ImageNet-C dataset. Different colors denote the values of $\alpha$, which regulates the strength of HIR loss w.r.t. the classification loss. AGG is the baseline setting with only a classification loss without using HIR loss.}
    \label{fig:tiny_hist}
\end{figure*}
The uniqueness of RMNIST dataset is that it contains paired images across domains which is the same image with different rotated angles. We investigate the influence of our HIR loss with paired and unpaired training batch. We set the batchsize to be 50 to contain one sample per domain per class. The samples from different domains are paired images in the paired setting and vice versa. The results are compared in Table~\ref{tab:pair_rmnist}. The baseline results of paired and unpaired settings are close while the HIR loss makes significant difference. This is because the posteriors of paired images are similar and further regularizing the HIR loss leads to overfit on each pair of images. The HIR losses of paired and unpaired experiments are visualized in Fig.~\ref{fig:kl_losses} for each domain. HIR works better in the unpaired setting where the divergences between posteriors are larger.

         

\subsubsection{\textbf{VLCS}}
Unlike rotated MNIST, VLCS is consisted of four independent datasets where the global structure of domains is not obvious. This experiment further demonstrates the effectiveness of HIR loss on datasets without any order in the domains. The number of samples varies across domains and classes and we do stratified sampling. Samples of each training domain are split into 80 folds with balanced classes in each fold. One training batch is consisted of a fold from each domain. We use Adam with a learning rate of 1e-4. The coefficient $\alpha$ is 1e-6 to match the scale of the empirical loss.

For VLCS, we use the CCSA architecture \cite{motiian2017unified} with two fully connected layers of dimension 1024 and 128. We adopt the same experiment setting as in CCSA where the dataset is randomly split into 0.7 for training and 0.3 for testing. Results are averaged across 20 repetitions. We initiate an individual random split for each repetition, which causes the relatively high standard deviations. The results of AGG and HIR are presented together with other existing approaches in Table~\ref{tab: vlcs}.

\subsubsection{\textbf{Tiny ImageNet-C}}
We show that data augmentation, especially when the augmented corruption types are divergent, fits well in a domain generalization setting. Our HIR loss can help with aligning the divergence between different corruptions.

Paired images of the 7 domains in Tiny ImageNet-C are visualized in Fig.~\ref{fig: corruptions} together with the uncorrupted image. We adopt ResNet50 pretrained on ImageNet for this experiment, so the uncorrupted images are not included as one domain in this setting. This dataset is challenging in the way that the corruptions are severe and both blurring and noise corruptions are presented. Training on one type of corruption cannot help but may deteriorate the performance on an unseen corruption.

We use the Adam optimizer with learning rate 1e-5. A small batch is consisted of one image and all its paired images from all the training domains and a large batch contains 20 shuffled small batches. The large batch is used as one batch during training. Unlike the rotated MNIST dataset, due to the large domain shifts in this dataset, even posteriors of paired images have large variation  so the HIR loss of paired images is high enough to contribute to the posterior alignment.  We show the impact of coefficient $\alpha$ for all the seven domains in Fig.~\ref{fig:tiny_hist}. For domains with noise corruptions, larger $\alpha$ works better while smaller $\alpha$ is more suitable for blurring corruptions.

\section{Conclusion}\label{section: conclusion}
This work summarizes existing approaches for domain generalization in probabilistic expressions and shows that learning DIRs is too strict for representation learning so useful domain information is discarded. We proposed to learn HIRs instead of DIRs aiming to keep possible global structure of the domains without prior knowledge of domains, thus the target domain can be inferred from the relation between domains. 

In our work, to avoid arbitrary density estimation of the posterior of each domain, we approximated it by aligning the posteriors of samples from each domain. Future work can explore how to reliably estimate the distribution of domain posteriors to further relax the invariance learning.

\section*{Acknowledgment}
This work is part of the research programme C2D–Horizontal Data Science for Evolving Content with project name DACCOMPLI and project number 628.011.002, which is (partly) financed by the Netherlands Organisation for Scientific Research (NWO).




\bibliographystyle{IEEEtran}
\bibliography{IEEEexample}
%


%

\end{document}